\documentclass[preprint,12pt,number,sort&compress]{elsarticle}

\usepackage{amsmath}
\usepackage{amssymb}
\usepackage{graphicx}
\usepackage{url}
\usepackage{soul}
\usepackage{caption}
\usepackage{siunitx}
\usepackage{multirow}
\usepackage{enumitem}
\usepackage{float}
\usepackage{lineno}

\usepackage{booktabs, multirow}
\usepackage[table]{xcolor}

\definecolor{blue}{HTML}{008ED7}
\definecolor{mygray}{gray}{0.75}
\definecolor{lightblue}{HTML}{e5f7ff}
\definecolor{darkgreen}{HTML}{009B55}

\newcommand{\tabletitle}[1]{\textbf{\textcolor{blue}{#1}}}
\newcommand{\tablesubtitle}[1]{\textcolor{blue}{#1}}

\begin{document}

\begin{frontmatter}

\newpage

\Large 

\noindent \textbf{Title:} Synthesizing Late-Stage Contrast Enhancement in Breast MRI: A Comprehensive Pipeline Leveraging Temporal Contrast Enhancement Dynamics 

\normalsize

\begin{itemize}
    \item  Ruben D. Fonnegra* \\
    \textbf{Affiliation:} Institución Universitaria Pascual Bravo, Instituto Tecnológico Metropolitano \\
    \textbf{City/Country:} Medellín, Colombia \\
    \textbf{Email:} ruben.fonnegra@pascualbravo.edu.co \\
    \textbf{Permanent email address:} rubenfonnegrat1015@gmail.com  \\

    \item Maria Liliana Hernández \\
    \textbf{Affiliation:} Ayudas Diagnósticas SURA S.A.S. \\
    \textbf{City/Country:} Medellín, Colombia \\

    \item Juan C. Caicedo \\
    \textbf{Affiliation:} Morgridge Institute for Research, Dept. Biostatistics and Medical Informatics, University of Wisconsin—Madison \\
    \textbf{City/Country:} Madison, WI, USA \\

    \item Gloria M. Díaz \\
    \textbf{Affiliation:} Instituto Tecnológico Metropolitano \\
    \textbf{City/Country:} Medellín, Colombia \\
\end{itemize}

\newpage


\title{Synthesizing Late-Stage Contrast Enhancement in Breast MRI: A Comprehensive Pipeline Leveraging Temporal Contrast Enhancement Dynamics}

\author[O1,O2]{Ruben D. Fonnegra\corref{cor1}} \ead{ruben.fonnegra@pascualbravo.edu.co} 
\author[O3]   {Maria Liliana Hernández} 
\author[O4]   {Juan C. Caicedo} 
\author[O2]   {Gloria M. Díaz} 

\cortext[cor1]{Corresponding author}

\affiliation[O1]{organization={Institución Universitaria Pascual Bravo},
            city={Medellín},
            country={Colombia}}

\affiliation[O2]{organization={Instituto Tecnológico Metropolitano},
            city={Medellín},
            country={Colombia}}

\affiliation[O3]{organization={Ayudas Diagnósticas SURA S.A.S.},
            city={Medellín},
            country={Colombia}}

\affiliation[O4]{organization={Morgridge Institute for Research, Dept. Biostatistics and Medical Informatics, University of Wisconsin—Madison},
            city={Madison},
            state={WI},
            country={USA}}

    
\begin{abstract} 
Dynamic contrast-enhanced magnetic resonance imaging (DCE-MRI) is essential for breast cancer diagnosis due to its ability to characterize tissue based on contrast agent kinetics. conventional DCE-MRI protocols require multiple imaging phases, including both early and late post-contrast acquisitions, leading to prolonged scanning times that can cause patient discomfort and  motion artifacts, as well as contribute to higher costs and limited availability in clinical settings. To address these limitations, this paper presents a comprehensive pipeline for synthesizing long-term (late-phase) contrast-enhanced breast MRI images from short-term (early-phase) counterparts, aiming to replicate the behavior of the time-intensity (TI) curve in enhanced regions while maintaining visual properties across the entire image. The proposed approach introduces a new loss function called the Time Intensity Loss (TI-loss), which leverages the temporal behavior of the contrast agent to guide the training of a generative model. Furthermore, as established normalization strategies shown undiserable effects on the enhancement beahviour, a novel normalization strategy (TI-norm) is also proposed, which preserve the contrast enhancement pattern across multiple image sequences at various timestamp. Additionally, two new metrics are proposed to evaluate the synthesized image quality, i.e., the Contrast Agent pattern score ($\mathcal{CP}_{s} $), which determines the validity of annotated regions according to their enhancement patterns (plateau, persistent, washout), and the average difference in enhancement ($\mathcal{ED}$), that quantify the difference between the real and generated enhancement in selected regions. Evaluation was performed using public DCE-MRI dataset that includes studies from 3T and 1.5T scanners with different imaging techniques. Experimental results demonstrate that our method accurately synthesizes the contrast enhancement response in terms of the TI curve in regions of interest, that significantly outperforms other models, while maintaining visual properties comparable to real late-phase contrast-enhanced images. By enabling accurate synthesis of late-phase contrast-enhanced images from early-phase data, our method has the potential to optimize DCE-MRI protocols, reducing scanning time without compromising diagnostic accuracy. This advancement brings generative models closer to practical implementation in clinical scenarios, enhancing efficiency in breast cancer imaging.
\end{abstract}


\begin{graphicalabstract}
\includegraphics[width=\textwidth]{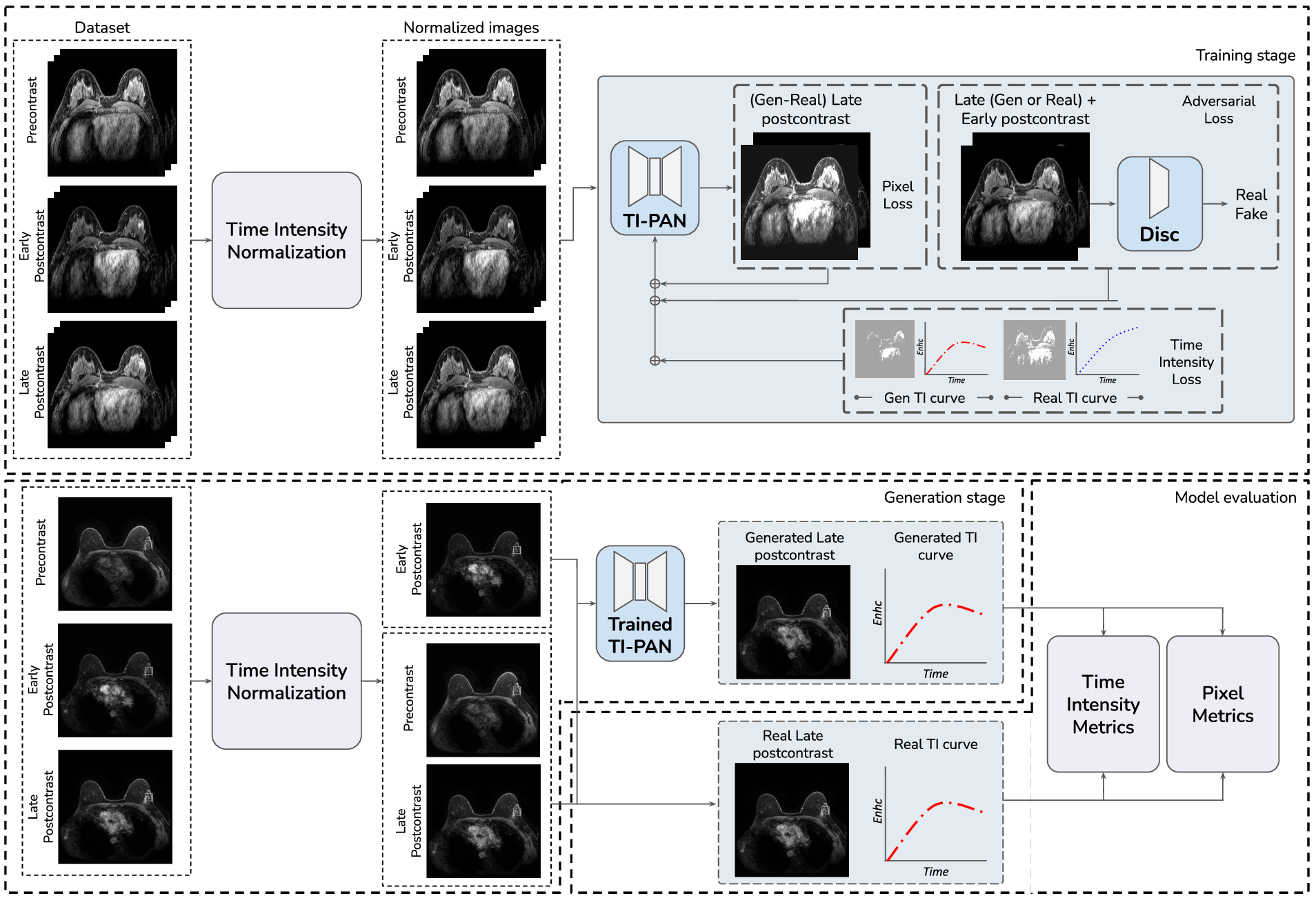}
\end{graphicalabstract}


\begin{highlights}
    \item A comprehensive pipeline for synthesizing late-phase contrast-enhanced breast MRI images from early-phase that ensures coherent image synthesis and accurate clinical interpretation, bringing generative models closer to practical implementation in clinical scenarios.
       
    \item A new loss function (TI-Loss) that leverages the temporal behavior of contrast agents to guide the training of generative models, ensuring accurate replication of contrast enhancement patterns.
    
    \item A novel normalization strategy (TI-norm) for DCE-MRI imaging that preserves the kinetic integrity of the time-intensity curve across multiple image sequences and timestamps showing a critical effect during model training.
    
    \item Two new TI curve-based metrics to objectively assess the quality and clinical reliability of generated images according to their real biological behavior.
\end{highlights}


    

\makeatletter
\def\ps@pprintTitle{%
  \let\@oddhead\@empty
  \let\@evenhead\@empty
  \def\@oddfoot{\reset@font\hfil\thepage\hfil}
  \let\@evenfoot\@oddfoot
}
\makeatother

\end{frontmatter}



\section{Introduction} \label{sec:introduction}

Breast cancer remains one of the leading causes of morbidity and mortality among women worldwide, and early detection and accurate diagnosis are critical for improving prognosis and reducing the mortality rate associated with it ~\cite{Miller2022Cancer,Bhushan2021Current}. Dynamic contrast-enhanced magnetic resonance imaging (DCE-MRI) has become an essential tool in evaluating breast lesions due to its high sensitivity and capability to characterize breast tissue and the internal structures of the tumor ~\cite{mann2019breast,potsch2022Contrast}. This is made possible by the intravenous administration of a gadolinium-based compound known as a contrast agent (CA). Due to the accelerated metabolic activity of malignant tumors, they exhibit a characteristic CA absorption pattern over time that differs from benign and healthy tissues in both speed and intensity. This pattern is highly sensitive and facilitates tumor detection and characterization, making it easier in comparison to other imaging modalities. For radiological findings (i.e. suspicious lesions), the CA absorption pattern is defined by the analysis of a curve known as the kinetic or time-intensity curve (TI), which is computed from the variation of pixel intensity at the location of the maximum intensity point from early or short-term (1-3 min) and late or long-term (8-10 min) responses. This analysis, along with the visualization of the internal structure of tumors, makes DCE-MRI one of the most efficient tools for detecting and characterizing breast cancer ~\cite{Rizzo2021Preoperative}.

Although DCE-MRI protocols vary even from hospital to hospital, conventionally, they involve acquiring multiple sequences after CA administration, including at least both early and late contrast response images, which significantly prolongs the scanning time. This extended duration can cause discomfort for patients, increase the risk of motion artifacts, and limit the availability of MRI equipment for other examinations.  Furthermore, the large number of images per patient requires substantial infrastructure in terms of storage and transmission devices in the database center and the picture archiving and communication system (PACS). Consequently, radiologists must spend a considerable amount of time interpreting images to reach a diagnostic conclusion, given the volume of information per patient to analyze. Alternatively, ultra-fast DCE-MRI (UF-DCE-MRI) has emerged to capture kinetic information in the very early post-contrast period by drastically increasing the temporal resolution, which significantly reduces acquisition time. However, increasing temporal resolution also decreases spatial resolution, leading to inaccuracies in the visualization of internal tissue structures ~\cite{kataoka2022ultrafast}. As another option, abbreviated MRI protocols have been proposed to reduce the number of image sequences that must be acquired, stored, and interpreted ~\cite{kuhl2014abbreviated,Hernandez2021a}. Typically, these protocols include only one early post-contrast image sequence, which allows for the detection of image findings or lesions but limits tumor characterization because they do not capture the dynamic time-intensity (TI) pattern that distinguishes between probably benign and probably malignant enhanced masses ~\cite{hernandez2021magnetic}. As an example, biologically aggressive cancers will exhibit strong angiogenic activity with fast wash-in and wash-out, whereas fibroadenomas often exhibit fast wash-in, although wash-out is scarce ~\cite{Kuhl2024Abbreviated}. 

Recently, generative artificial intelligence has shown the potential to synthesize images, and numerous studies have demonstrated the feasibility of using these models in several medical imaging domains~\cite{wang2021synthesizing,conte2021generative,ozbey2023unsupervised}. Particularly, synthesis of contrast-enhanced images has been proposed as an alternative to reducing or eliminating the need for contrast agents~\cite{luo2021deep,Muller-Franzes2023Using}. In the case of breast DCE-MRI, sSome studies have explored generating virtual contrast images from non-contrast MRI scans, aiming to eliminate the need for contrast agents altogether~\cite{haase2023reduction}. Consequently, generative models incorporating specific pixel-based metrics or morphological features have been presented to surpass precise local tumor visualization~\cite{haarburger2019multiparametric,Kim2021Generative,kim2022tumor,Xie2022Magnetic}. Nonetheless, these works have been constrained by their limited ability to furnish relevant information that extends beyond their realism for diagnostic purposes~\cite{liebert2024impact}. 

Initial studies focused on generating the first post-contrast sequence from pre-contrast or non-contrasted images through the use of generative artificial intelligence methods ~\cite{haarburger2019multiparametric,Kim2021Generative,rincon2021analysis}. Furthermore, to enhance image quality in the visualization domain, subsequent studies have suggested incorporating additional visual cues, such as morphological or shape features. ~\cite{kim2022tumor,Xie2022Magnetic,canaveral2024postcontrast}. Simultaneously, other authors have assessed the utilization of different image sequences, including diffusion-weighted and apparent diffusion coefficient images, regarding their influence on the visual domain of synthesized images~\cite{zhang2023synthesis,ramanarayanan2024dce}, and T2-weighted images~\cite{muller2023using,liebert2024feasibility}. Recently, a comprehensive evaluation of several input sequences for generating virtual contrast-enhanced images showed that, even though quantitative metrics indicate similar performance in almost all cases, there are significant differences in qualitative evaluations regarding qualitative scores, including diagnostic image quality, image sharpness, satisfaction with image contrast, and visual signal-to-noise ratio ~\cite{liebert2024impact}. Thus, despite the fact that Although these methods hold significant technological promise, their current lack of clinical confidence and inability to capture essential contrast dynamics necessary for accurate lesion characterization cast doubt on their near-term applicability in clinical practice.

In order to address the diagnostic need for temporal evaluation of lesion dynamics and obtain a comprehensive and useful virtual CA response for diagnosis, a few works have proposed synthesizing the late-phase contrast image. Oh et al.~\cite{oh2022tdm} proposed generating a "late-phase" (210s after CA administration) from the early UF-DCE-MRI sequence (70 and 140s after CA administration) using a modified paired StarGAN model and the difference phase map to specifically focus on tumor information during training. The approach's results are promising; however, it is based on UF-DCE-MRI, which has low spatial resolution. Furthermore, generated images are equivalent to early post-contrast in conventional DCE-MRI and may not capture the required long-term information. Li et al.~\cite{li2024mtfn} introduced a multi-temporal fusion network that fuses low-dimensional attention maps from the first and third (early) post-contrast phase images to generate the eighth phase image (late). Feature fusion is performed using a pixel-level information co-attention model that attempts to highlight enhancement regions where a contrast agent response could be found. While this study has effectively used temporal image sequences to predict late responses, it still relies on a pixel-only technique, which could fail to provide assessments that are both interpretable and clinically pertinent. This gap underscores the need for strategies and evaluation frameworks that align with the diagnostic tools and interpretative practices widely accepted in the radiological community. 

Conversely, we propose attempts to capture the dynamic behavior of the contrast agent and use this to train generative model frameworks that synthesize late-phase images, preserving this behavior while maintaining spatial characteristics throughout the entire image. In our prior research~\cite{Fonnegra2023early}, we presented a loss function that approximates the expected reaction response of the contrast agent in tissue by reducing the overall difference between the real and predicted enhancement patterns. The results of this study motivated the exploration of generative models aimed at emulating the clinical information response within a generative imaging framework. 

 Since the CA pattern of the TI curve is the most relevant information to distinguish benign from malignant lesions in DCE-MRI, the synthesis of contrast enhancement images must accurately simulate this pattern rather than synthesize a realistic image only. With this aim, this study presents a robust and comprehensive pipeline for creating long-term (late) contrast-enhanced breast MRI images from their short-term (early) counterparts. This method effectively models the TI curve behavior in enhanced regions while maintaining visual and structural integrity throughout the image, thereby addressing significant deficiencies in clinical imaging workflows and establishing a new standard for the production of high-quality, diagnostically pertinent MRI outputs. This pipeline intends to synthesize images that preserve accurate interpretability to approach generative models for their implementation in clinical scenarios.

To achieve this, specific components involved in each processing step were addressed. In the same way as our previous work ~\cite{Fonnegra2023early}, we proposed a new loss function named the Time Intensity Loss function (TI-loss) that takes advantage of the temporal behavior of the CA to guide the training of a generative model to synthesize images that preserve both the CA pattern and spatial features of the generated image. In addition, unlike traditional normalization and standardization methods, we developed a novel strategy that maintains the contrast enhancement pattern across several image sequences at various timestamps. Finally, since the quality of the synthesis process depends on the ability of the model to predict the CA pattern, we propose an evaluation based on two new metrics, the CA pattern score ($\mathcal{CP}_s$) that determines the validity of the annotated regions according to their CA pattern (plateau, persistent, and washout), and the average difference in the enhancement ($\mathcal{ED}$), which quantifies the difference between the real and generated enhancement in unannotated regions within the tissue. Pipeline performance was evaluated using a public DCE-MRI image dataset that included studies from 3T and 1.5T scanners with different imaging techniques ~\cite{saha2018machine}. 

The experimental results show that the proposed pipeline generates images that accurately synthesize the contrast enhancement response in terms of the TI curve for both radiological findings (regions of interest - ROI) and unannotated regions, while maintaining visual properties comparable to the real late contrast-enhanced images. This result,  achieved primarily by implementing the TI-Loss during generative model training, is extremely important, as the evaluation of the pixel-only models indicates that optimizing for pixel-wise quality does not guarantee the prevalence of clinically relevant information. However, the use of TI-Loss alone does not achieve optimal results in terms of clinical performance. The role of the TI-norm was evaluated, demonstrating the prevalence of clinical interpretation and behavior among images to facilitate learning of their relationships while maintaining visual quality in the images. The comparative analysis of the models shows that the proposed pipeline allows for the generation of images that preserve spatial properties in terms of pixel metrics, while significantly outperforming them in clinical interpretation.

\section{Methods}\label{sec_method}

\begin{figure*}
    \centering
    \includegraphics[width=\textwidth]{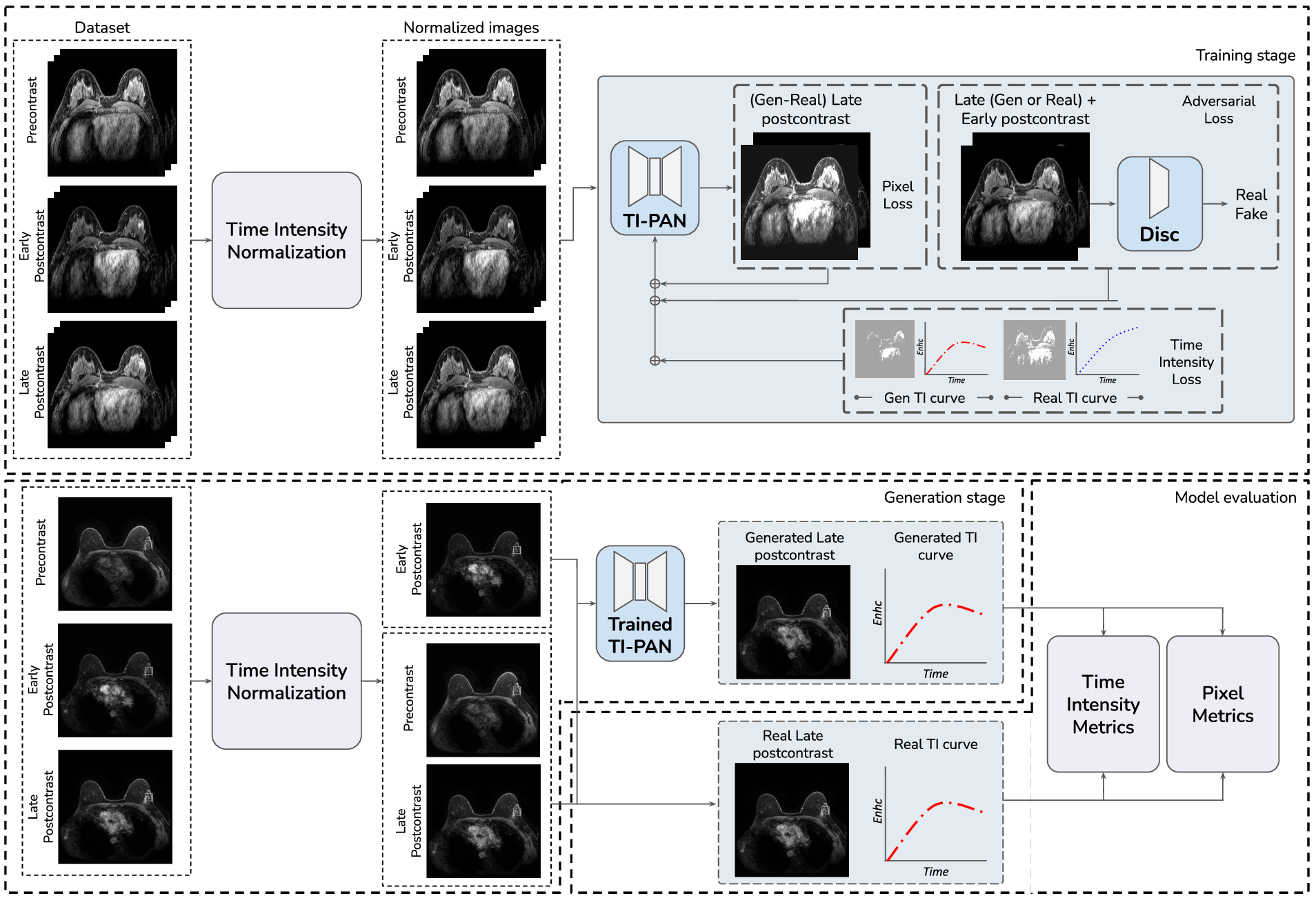}
    \caption{Graphical representation of our proposed pipeline. It is composed of three main stages: Generative model training, late post-contrast generation and model evaluation. For pre-processing, we proposed the Time Intensity normalization (TI-norm) to ensure the prevalence of the CA behavior among sequences. In the train stage, we proposed the Time Intensity Loss (TI-Loss) to leverage the contrast-enhancement behavior to outperform the diagnostic value of the generated images. The generation stage allows the synthesis of post-contrast images, and for the evaluation stage we proposed a set of metrics based on the TI curve.} \label{fig_graph_abs}
\end{figure*}

A graphical representation of the proposed pipeline is presented in Figure \ref{fig_graph_abs}. It is composed of three main stages: generative model training, late post-contrast synthesis, and model evaluation. Before training the generative model, training images are normalized using a novel normalization technique named time-intensity normalization (TI-norm), which attempts to avoid losing the intensity changes among dynamic sequences, i.e. pre-contrast and post-contrast images. Then, a Pixel Attention Network (PAN) is used as the backbone to synthesize the late post-contrast image from the early counterpart. A novel loss function called TI-loss is proposed here to guide the learning process to minimize the difference between the real and late CA responses. The trained PAN using the TI-loss (TI-PAN) is then used to generate a synthetic late post-contrast image from the TI-normalized early image. Finally, the performance of the late contrast synthesis is evaluated in twofold. First, conventional image quality metrics (MAE, SSIM and PSNR) between real and generated late post-contrast images are computed to determine the spatial quality of the generated images. Second, two metrics are used to determine the difference between the real and generated time-intensity curves, allowing for the evaluation of the ability to maintain the clinical significance of the CA pattern in the synthesized image.

\subsection{Time Intensity Pixel Attention Network: TI-PAN}

As mentioned above, the proposed pipeline is based on synthesizing the CA pattern in the best way. We used an attention-based model called the Pixel AAttention Ntwork (PAN)~\cite{zhao2020efficient}. This model employs attention layers under a unique scheme of blocks known as self-calibrated convolution (SC-PA) for feature extraction and nonlinear mapping, and nearest-neighbor upsampling (U-PA) for reconstruction. These blocks are considerably more effective than their conventional residual/dense counterparts at a lower parameter cost. Once the model receives an early-response input image $x_e$, the PAN extracts features at the spatial level and stacks multiple SC-PA blocks to generate powerful representations. Subsequently, U-PA blocks are used as reconstruction modules to upsample the features. After reconstruction, a global skip connection path is employed, wherein a bilinear interpolation to the input is performed. Then, the output corresponding to the late response $y_{l'}$ is obtained. Finally, to encourage realism in the image, we added a discriminator network and incorporated an adversarial approach, similar to that used in generative adversarial networks (GANs) \ref{eq_gan}. This also includes the pixel-wise loss based on the $L1$ norm to enforce spatial dependency. The full objective for the PAN model is shown in equation \ref{eq_full_obj}

\vspace{-0.4cm}

\begin{eqnarray} \label{eq_gan}
\begin{array}{ll}
    \mathcal{L}_{GAN} (G, D) = & \mathbb{E}_{X_e, Y_l} [log D(x_e, y_l)] + \\ 
    & \mathbb{E}_{X_e} [log(1 - D(x_e, G(x_e))] 
\end{array}
\end{eqnarray}

\begin{equation} \label{eq_full_obj}
    G^* = \mathrm{arg} \ \underset{G}{min} \ \underset{D}{max} (G, D) \ \mathcal{L}_{GAN} + \ \lambda_{I} \mathcal{L}_{I} 
\end{equation}

\subsubsection*{Time-Intensity Loss for learning late CA response}

To help generative models understand contrast agent behavior in tissue, we aim to formulate a general function that models the behavior of the CA response in the images. To achieve this, we created a loss term to determine the difference in the enhanced areas by the contrast agent, called the Contrast Enhancement Loss (CELoss) ~\cite{Fonnegra2023early}. Specifically, we consider the real early ($x_e$) and late ($y_l$) responses sampled from data $X_e$ and $Y_l$ respectively; we compute the real contrast-enhanced map ($CE_r(x_e, y_l)$) by subtracting the early response from the late response (equation \ref{eq_r_enhmap}) using the $L1$ norm. In a similar manner, the contrast-enhanced map is also computed for the generated image $CE_g$ (equation \ref{eq_g_enhmap}). The model is trained to minimize the difference between $CE_{r}$ and $CE_g$ using the $L1$ norm as formulated in equation \ref{eq_celoss}.

\vspace{-0.3cm}

\begin{equation} \label{eq_r_enhmap}
    CE_r(x_e, y_l) = \mathbb{E}_{X_e, Y_l} \left [ \ \left \| y_l - x_e  \right \| _1 \ \right ]
\end{equation}


\begin{equation} \label{eq_g_enhmap}
    CE_g(x_e) = \mathbb{E}_{X_e} \left [ \ \left \| G(x_e) - x_e \right \| _1 \ \right ]
\end{equation}


\begin{equation} \label{eq_celoss}
    \mathcal{L}_{CE} (x_e, y_l) =  \left [ \ \left \| CE_r(x_e, y_l)  - CE_g(x_e)  \right \| _1 \ \right ]
\end{equation}

Despite its nice performance and convergence, this method is susceptible to the existing class imbalance between the amount of background and biological signal pixels. This causes gradient diminishing and suboptimal solutions, especially when the $CE$ value is reduced after a few training epochs. To overcome this limitation, we introduce a subtle but important change by replacing the $L1$ with a more flexible function: the Huber loss ~\cite{Huber1964Robust}. The Huber loss uses $L1$ and $L2$ in the estimate to handle outliers, preserves smooth differentiability, and balances accuracy and generalization. It behaves as a quadratic function ($L2$) for values below $\lambda$ and as a linear function ($L1$) for greater values. The formulation of the Huber loss, $\mathcal{L}_{TI} (x_e, y_l)$, applied to the contrast enhancement problem is shown in equation \ref{eq_cmloss}, where $CE_r$ and $CE_g$ are functions of $(x_e, y_l)$ and $(x_e)$ respectively, but simplified in notation. We rename this function as the Time-Intensity Loss (TI-Loss):

\begin{equation} \label{eq_cmloss}
    \mathcal{L}_{TI} = \\
    \left\{
    \begin{array}{ll}
    1/2 \cdot [CE_r - CE_g] ^2, \ when \ \| CE_r - CE_g \| \leq \delta \\ 
    \delta \cdot [ \| CE_r - CE_g \| - 1/2 \cdot \delta ] \ otherwise.
    \end{array}\right.
\end{equation}


\begin{equation} \label{eq_full_obj}
    G^* = \mathrm{arg} \ \underset{G}{min} \ \underset{D}{max} (G, D) \ \mathcal{L}_{GAN} + \ \lambda_{I} \mathcal{L}_{I} + \ \lambda_{TI} \mathcal{L}_{TI}
\end{equation}

Hence, combining the TI-Loss with the adversarial approach of the PAN can be defined as shown in equation \ref{eq_full_obj}, where $\lambda$ is a weighting hyperparameter for the loss function.

\subsection{Image normalization based on the Time-Intensity pattern} \label{sec_data_norm}

Image normalization plays an important role in image synthesis. Most related works use conventional normalization techniques such min-max and z-score normalization. However, normalizing pixel intensity based on single-image statistics can distort diagnostic information because it varies temporally with other precontrast and post-contrast sequences. For this reason, we designed a normalization strategy to ensure that the information in the TI curve is preserved across image sequences. This technique is based on the Z-score standardization with respect to a different reference. For a given precontrast image $x_p$, both $\mu$ and $\sigma$ are computed as in the Z-score and the normalized image $x_{p_{TI}}$ is calculated. However, the early ($x_e$), real late ($y_l$) and generated late ($y_{l'}$) images are standardized using the $\mu$ and $\sigma$ values from $x_p$. This will ensure that pixel values for all the post-contrast images will reflect the variation in terms of intensity with respect to the precontrast to effectively preserve the contrast enhancement pattern. We named this as Time-Intensity nomralization (TI-norm). To compute the TI-norm of $x_p, x_e, y_l$ and $y_{l'}$, we use equations in \ref{eq_zscore_ref} 


\begin{eqnarray}
    \begin{array}{lll}
    x_{p_{TI}} = \frac{x_{p} - \mu_{x_p}}{\sigma_{x_p}}, & x_{e_{TI}} = \frac{x_{e} - \mu_{x_p}}{\sigma_{x_p}}, & 
    y_{l_{TI}} = \frac{y_l - \mu_{x_p}}{\sigma_{x_p}}, \\ \\ & y_{l'_{TI}} = \frac{y_{l'} - \mu_{x_p}}{\sigma_{x_p}} \label{eq_zscore_ref}
    \end{array}
\end{eqnarray}


\subsection{Quality evaluation based on Time-Intensity curve} 
\label{sec_ti_metrics}

Since the quality of our proposed approach depends on the ability to replicate the expected attributes of the time-intensity curve, we proposed two metrics based on its behavior, in addition to conventional pixel-based metrics. For a set of pre-contrast ($x_p$), early ($x_e$), late ($y_l$) and generated ($y_{l'}$) responses, a real and a synthetic time-intensity curve are computed, and three groups of metrics are estimated:

\subsubsection{CA pattern score in annotated regions ($\mathcal{CP}_{s} $)} This metric aims to determine if the CA pattern (plateau, persistent, and wash-out) in a region of the image with a small field of view is the same as expected. This approximation aligns with medical practice, where the TI curve is computed in a single or small valley of pixels to analyze its behavior across multiple image sequences. For a set of regions in all image modalities ($Rx_p$, $Rx_e$, $Ry_l$ or $Rx_p$, $Rx_e$, $Ry_{l'}$) the time intensity curve is computed. To determine the CA pattern for each of them, we calculate the percentage of enhancement ($\mathcal{E}_{ps}$) of the late response with respect to the early phase as shown in equation \ref{eq_enhancement}. Then, the type of TI curve is estimated following the standard rules shown in equation \ref{eq_curves}. Finally, the overall multi-class F1 score between them is computed and reported.

\begin{equation}
    \mathcal{E}_{ps}(Rx_e, Ry_l) = \frac{Ry_l - Rx_e}{Rx_e} \times 100 (\%) \label{eq_enhancement}
\end{equation}


\begin{equation}
    \mathcal{CP}_{s} =
\begin{array}{ll}
 Persistent, & \mathcal{E}_{ps} > +10\% \\
 Plateau,    & -10\% \leq \mathcal{E}_{ps} \leq +10\%  \\ 
 Washout,   & \mathcal{E}_{ps} <  -10
\%
\end{array}
\label{eq_curves}
\end{equation}

\subsubsection{Average difference of $\mathcal{E}_{ps}$ in regions ($\mathcal{ED}$)} The purpose of this metric is to estimate the difference between both real and generated TI responses within the tissue under different conditions. In this case, the $\mathcal{E}_{ps}$ is computed in pairs of real and generated regions to determine how different the expected score is synthesized. This difference is calculated instead of the CA pattern given its clinical significance, which is bound to clinically relevant findings. Formally, given a set of annotated ROIs ($Rx_p$, $Rx_e$, $Ry_l$ or $Rx_p$, $Rx_e$, $Ry_{l'}$) or a set of unannotated regions ($Px_p$, $Px_e$, $Py_l$ or $Px_p$, $Px_e$, $Py_{l'}$) in all image modalities, the TI curve is calculated, as well as the percentage of enhancement ($\mathcal{E}_{ps}$). Then, the $L1$ norm is used to calculate the distance between the real and generated TI curves at the late phase from the annotated ROIs ($\mathcal{ED}_{R}$) or unannotated regions ($\mathcal{ED}_{UR}$). The formulation of the score for both scores is shown in equation \ref{eq_enhancement}.

\begin{equation}
\begin{array}{c}
 \mathcal{ED}_{R}({ Rx_e, Ry_l, Ry_{l'} }) = \left\| { \mathcal{E}_{ps}(Rx_e, Ry_l) - \mathcal{E}_{ps}(Rx_e, Ry_{l'}) } \right\|  \\
 \mathcal{ED}_{UR}({Px_e, Py_l, Py_{l'} }) = \left\| { \mathcal{E}_{ps}(Px_e, Py_l) - \mathcal{E}_{ps}(Px_e, Py_{l'}) } \right\|  \\
\end{array}
\label{eq_enhancement}
\end{equation}

\subsubsection{Image quality metrics} The purpose of this set of metrics is to evaluate the quality of the images in terms of spatial distribution. For this, we computed the Mean Absolute Error (MAE) to estimate the pixel-to-pixel change between real and generated sequences via spectral fidelity, the Structural Similarity Index Measure (SSIM) to estimate the structural quality of visual perception from three perspectives: correlation loss, luminance distortion, and contrast distortion, and the Peak Signal-to-Noise Ratio (PSNR) to determine the ratio between the maximum possible value (power) of the images and the power of distorting noise that affects their quality. Finally, to better understand the spatial differences in terms of intensity among the real and generated images, we visualized the absolute difference among each pixel intensity value as a heatmap.

\section{Results}

\subsection{Dataset}

To perform all experiments, we used the Duke dataset ~\cite{saha2018machine}. This dataset includes 922 preoperative MRI imaging sequences of invasive breast cancer patients (precontrast and 3 or 4 postcontrast). The patients were between 21 and 89 years old and comprised seven racial and ethnic groups. Tumor size, nuclear grade, and hormone receptor status varied among patients. Ten scanner manufacturers were used to capture MRI images, and patients received three kinds of contrast agents. These constraints maintain heterogeneous parameters and prevent the data from being limited to a single MRI configuration. Additionally, expert radiologists annotated at least one region with lesions or tissue anomalies for each patient. 

Our framework was built using pre-contrast, the earliest post-contrast (2 mins after CA administration), and the latest sequence (6 mins after CA administration). In our experiments, 1.5T and 3T images were used independently to train models due to their different visual features regarding contrast enhancement. The 1.5T and 3T subsets included 393 and 389 patients, respectively. Due to the sparse tissue contrast agent response, full DICOM volumes were not used in either case. A frame-wise selection was performed based on the available annotated ROIs, as well as previous and subsequent images. Thus, for training and validation, $9,196$ and $1,043$ images were extracted from the 3T images, and $8,941$ and $1,012$ from 1.5T. Additionally, images were normalized using TI-norm, as described in section \ref{sec_data_norm}.

\subsection{Experimental setup}

Regarding the PAN model, we ran three tests using the same training configuration and compiled the findings. For the baseline, we adopted the original implementation consisting of 16 SC-PA blocks and convolutions with a kernel size of 40 for downsampling and 24 for upsampling. Additionally, the discriminator network consists of a $1 \times 1$ Markovian discriminator ~\cite{Isola2017Image} with convolutional blocks (Conv2d + InstanceNorm + LeakyReLU) following the architecture as in ~\cite{isola2018imagetoimage}. Furthermore, using the same configuration, we trained the model including either CELoss (CE-PAN) or TILoss (TI-PAN) and compiled results to quantify the impact of the losses on the models' performance.

To perform a comparative analysis considering the CA behavior, we evaluated the quality of the generated images at three different levels: full image, annotated ROIs, and unannotated regions. The analysis of the full image is based on conventional pixel-wise metrics to check their realism, as in other similar works. On the other hand, the evaluation of annotated ROIs is based on the quality of the CA response generated using the metrics described in Section \label{sec_ti_metrics}. As the annotations correspond to large portions of tissue, the entire region was not considered. Instead, a small set of pixels was taken ($3 \times 3$), where its center corresponds to the maximum intensity point within the ROI.

However, since the expected behavior in annotated ROIs could bias the analysis in terms of variability, we also analyzed the generated CA response in tissue regions without annotated clinical findings. For this, regions within the tissue portion of the organs were manually selected in $x_e$, and their locations were projected in the rest of the modalities to compile $Px_p$, $Px_e$, $Py_l$, $Py_{l'}$. From each region, the TI curve and the increase in enhancement are computed as described in section \ref{sec_ti_metrics}. Additionally, to increase the number of samples to analyze, three(3) unannotated regions are extracted from each image in the 3T test set, to complete $3,129$ unannotated regions. At this point, the type of TI curve was not considered, as this characterization is usually assigned to lesions or clinically relevant findings. Instead, we expect no significant difference between the generated and expected increase in enhancement.

\subsection{Clinical Performance based on TI curve metrics} \label{sec12}

\begin{figure*}[t]
    \centering
    \includegraphics[width=\textwidth]{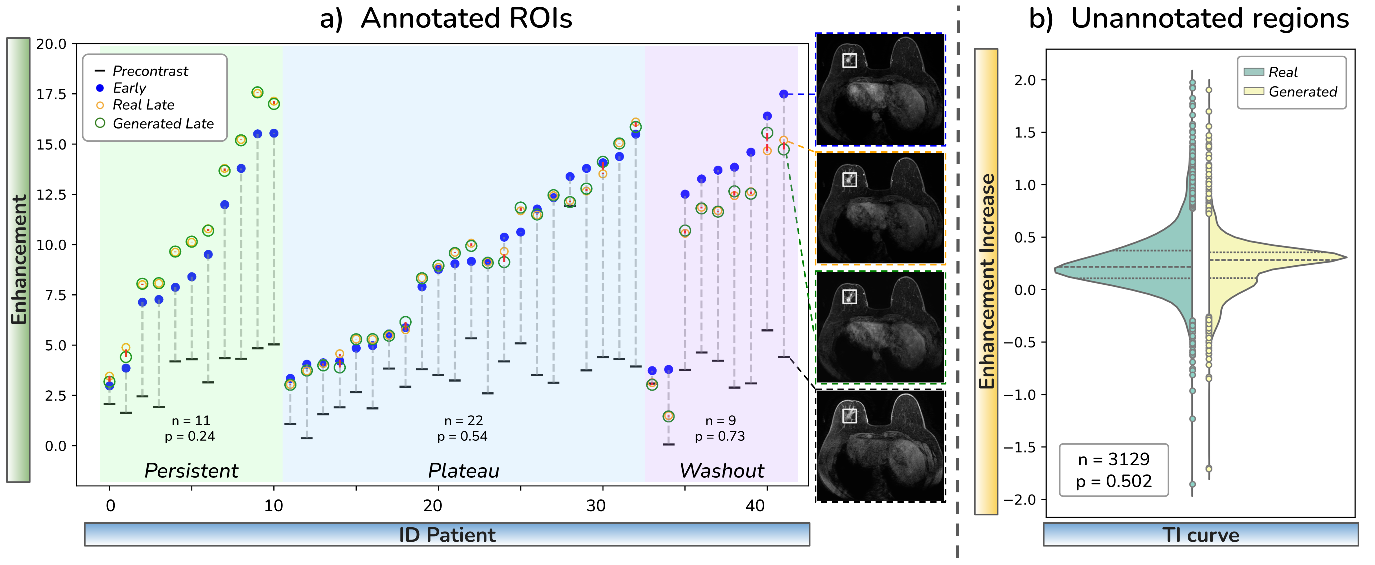}
    \caption{Results for TI curve estimation using our proposed pipeline. Figure a) shows the per-patient TI curve, where black lines and blue points correspond to the reference for the TI curve (pre-contrast and early post-contrast), orange points are the real CA response (late) and green points are the generated response. Additionally, the red dashed line shows the error distance between real and generated enhancement. Patients are organized according to their expected CA pattern (persistent, plateau and wash-out) and their early post-contrast enhancement in ascendingly. Figure b) shows the distribution of the enhancement increase/decrease among real and generated unannotated regions. Besides, dots in the tails of the distributions represent the outliers in terms of the change in increase. }
    \label{fig_res_clin}
\end{figure*}

Here, we evaluate the accuracy of the generated images for clinical applications. To achieve this, we used our proposed pipeline to predict the late response images given the early response. Then, the diagnostic value and performance in terms of the TI curve for the annotated ROIs and unannotated regions are analyzed using the aforementioned metrics. Figure \ref{fig_res_clin} a) shows the results for the model predicting the time-intensity response using TI-PAN with the 3T images, where patients are split according to their real CA pattern (persistent, plateau, and washout). The results show that the estimated CA pattern score for the annotated ROIs is $0.904$ and the expected TI response matches the real response for most patients, with small variations among them. 

Additionally, in the most prominent error cases, the proposed pipeline avoids the overestimation of the late CA response, which is important to avert false positive and false negative cases. Besides, in an in-depth analysis, only 5 cases were misestimated for the CA pattern. However, in all these cases, the average difference between the real and generated enhancement was $\pm 2.5\%$, and the real enhancement was near the threshold that determines the type of TI curve ($ \sim \pm 10\%$). Moreover, all cases were either real persistent or wash-out that were estimated as plateau, which indicates that they are not conclusive and require further examination. These findings are relevant for our purpose since they demonstrate the ability of the model to replicate the expected response in the TI curve for diagnosis reliably, while preserving the most important information for clinical analysis and avoiding false positive and false negative cases.

Finally, we aim to determine if there is a significant difference in the CA response between the real and generated groups of regions, according to their enhancement increase, by using the non-parametric Wilcoxon signed-rank test, where we chose a significance value of $0.05$. In the test, if the $p$-value rejects the null hypothesis, that means there is a significant difference between the real and generated CA responses. Otherwise, if the $p$-value fails the test, it means that there is no evidence to support a difference between real and generated CA responses. For the annotated ROIs reported in figure a), it can be observed that the $p$-values for the persistent, plateau, and wash-out TI curves are not able to reject the null hypothesis in any case, which supports that the model is able to generate the CA response following the same real pattern with no significant difference.

However, the analysis on the real annotated ROIs alone exhibits a limited number of samples and cases to analyze, which might bias the evaluation of the quality of the synthesized CA response in terms of generalization. For this reason, we extended the evaluation using unannotated regions within the tissue zones to highlight the changes. Unlike the previous analysis, the determination of the kind of TI curve is not made at this stage due to its clinical significance, which is only addressed for relevant findings. Instead, the analysis is performed in terms of the expected increase in enhancement to avoid misinterpretations. Plot b) in figure \ref{fig_res_clin} shows the distribution of the enhancement variation for the unannotated real and generated regions and its level of statistical significance using the aforementioned test. It can be observed that our pipeline is able to replicate the distribution of the increase in enhancement efficiently, with small differences around the mean. This is also demonstrated in the $p$-value obtained through the test, which indicates no evidence to support a significant difference in the increase in enhancement. Additionally, it is worth noting that outliers for both distributions are located below $-0.25 \%$ increase and above $+0.60 \%$ increase. These thresholds are important since they indicate that the most susceptible cases of analysis correspond to regions where higher or lower enhancement increases must be examined by clinicians, which correlates correctly with the characterization of wash-out and persistent TI curves. For the within-the-distribution cases, the results show that our pipeline is able to replicate the expected increase in enhancement with small variations.

\subsection{Impact of image normalization in the TI-Curve and CA pattern}

\begin{figure*}[t]
    \centering
    \includegraphics[width=\textwidth]{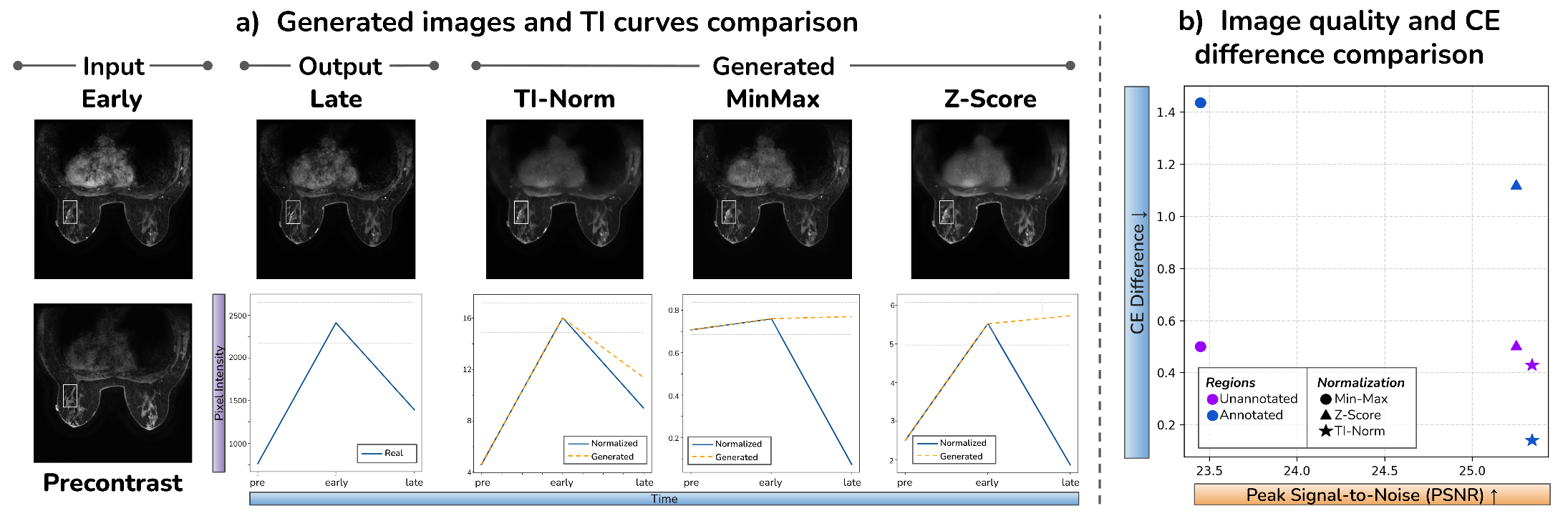}
    \caption{Impact of normalization in the diagnostic information. a) Top row displays the early (input) and late (output) responses, besides the generated images using the TI-PAN model. Bottom row shows the time intensity plots for annotated enhanced ROI, where blue continuous line corresponds to the computed real time-intensity curve in late and after normalization, and orange dashed line is the generated time-intensity curve by the model. b) Shows a comparative plot of image quality (PSNR) and CE difference using the model and different normalization strategies in annotated ROIs and unannotated regions.}
    \label{fig_res_norm}
\end{figure*}

At this point, we trained TI-PAN using Min-Max, Z-score, and TI-norm, and generated the synthetic images. In Figure \ref{fig_res_norm}, first row of a) shows the generated images from all models, where no major visual difference is evident. However, to assess the impact of the TI-norm on the diagnostic quality of the images, we computed the time-intensity curve of each annotated set of samples $Rx_p, Rx_e$ $Ry_l$ and $Ry_{l'}$. Using RadiAnt software ~\cite{radiant}, we computed the TI curve for the images without any normalization and after normalization. Figure \ref{fig_res_norm} displays time-intensity plots in the second row, where it can be seen that the TI curve is altered when conventional Min-Max and Z-score are used, in comparison to the non-normalized one. Nevertheless, the TI-PAN model still tries to mimic the given behavior when using Min-Max or Z-score with certain similarity. This is a critical issue for diagnosis since the information of the curve itself is degraded significantly, and it distorts its interpretability afterwards, despite the high visual correlation among real and generated late post-contrast images. Nonetheless, our proposed time-intensity normalization effectively preserves the CA behavior in the real and generated images, which approaches the generated TI curve to a real clinical interpretation.

We also extended the evaluation by comparing the predicted contrast enhancement difference and the image quality in annotated and unannotated 3T images, and the results are shown in b). Although, in this case, the Z-score norm obtains the best performance in terms of image quality, it falls very short in terms of the expected estimation of the CA pattern, achieving suboptimal performance. Similar behavior is also obtained using the Min-Max norm, where pixel error is lower but diagnostic quality remains suboptimal. In both cases, the TI-PAN model shows a decrease in performance and exhibits a critical limitation in the clinical scenario, despite using TI-Loss. In the case of TI-norm, it guarantees the prevalence of the TI curve to be leveraged by the TI-Loss during training. These features allow obtaining the best scores in the estimation of the CA pattern while achieving similar image-based quality (25.4 in PSNR). This makes TI-norm an important component in generating high-quality diagnostic information in an image synthesis setting, which has not been reported before in the medical imaging synthesis field.

\subsection{Quantitative and Qualitative evaluation of synthesized Images}

\begin{figure}[t]
    \centering
    \includegraphics[width=\textwidth]{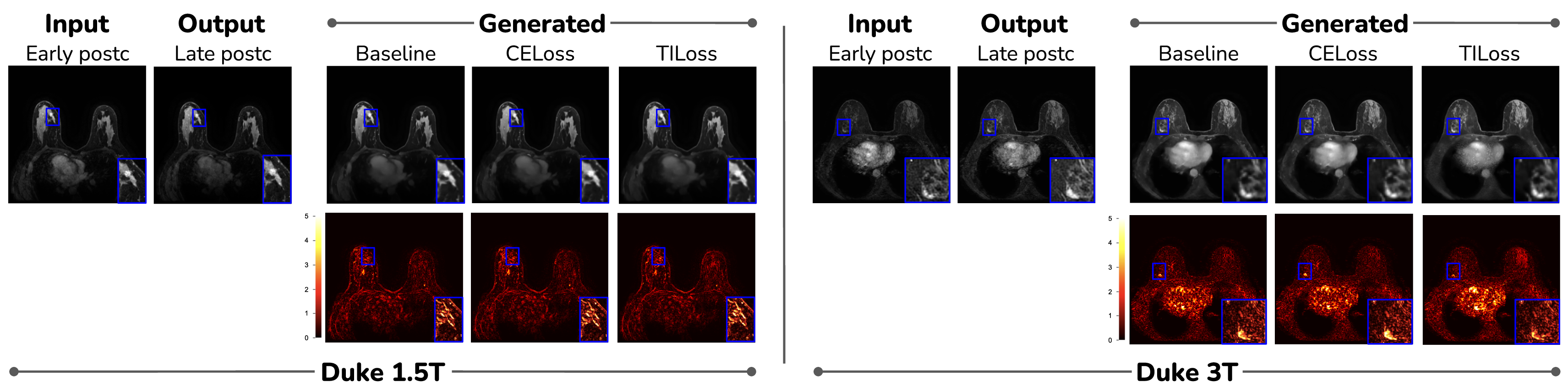}
    \caption{Comparative plots for visual image quality. Heatmaps represents the difference of the real vs the generated images in the same scale. Annotated ROIs are also shown in augmented projection. Best seen zoomed in.}
    \label{fig_comp_res}
\end{figure}

An evaluation of the visual quality of all the generated images at the spatial feature level was performed. Figure \ref{fig_comp_res} shows some generated sample results for images in the test set using all of the evaluated models together. Additionally, as images in the test set had at least one annotated ROI, they were also compared visually. Finally, the difference heatmaps provide compelling visual evidence of the pipeline's efficacy. For 1.5T images, the generated samples exhibit enough realism in the visual domain for nearly all models, with only minor deviations observed in specific tissue textures and particular details. This is coherent with the previous results where a narrow margin of difference was observed in the visual domain. However, in case of the 3T images, a smaller difference between the real and generated images is evidenced when using the TILoss, especially in the annotated region where the difference in the brighter zone is lower. Besides, table \ref{tab_comp1} also reports the MAE, SSIM and PSNR for all models at full image and ROI leves, where no single model obtains the best scores consistently. In contrast, no large margin is evidenced among models, and slight differences are found. This result is in line with other cutting-edge studies that have shown good results in creating useful images from post-contrast sequences ~\cite{Kim2021Generative,wang2021synthesizing}.

\subsection{Models Comparison}

\begin{table}[t]\centering
\begin{tabular}{lrrrrrrrr}\toprule
& &\multicolumn{3}{c}{\tabletitle{Duke 1.5T}} &\multicolumn{3}{c}{\tabletitle{Duke 3T}} \\\cmidrule{3-8}
& &\tablesubtitle{PAN} &\tablesubtitle{CE-PAN} &\tablesubtitle{TI-PAN} &\tablesubtitle{PAN} &\tablesubtitle{CE-PAN} &\tablesubtitle{TI-PAN} \\\cmidrule{3-8}
\multirow{3}{*}{\rotatebox[origin=c]{90}{\tabletitle{\footnotesize{\parbox{1.3cm}{ Pixel \\ (images)}}}}} &MAE &0.555 &0.554 &\textbf{0.543} &0.706 &\textbf{0.701} &0.736 \\
&PSNR &27.05 &26.96 &\textbf{27.08} &25.64 &\textbf{25.71} &25.54 \\
&SSIM &\textbf{0.689} &0.688 &0.687 &0.691 &\textbf{0.696} &0.688 \\\cmidrule{1-8}
\multirow{3}{*}{\rotatebox[origin=c]{90}{\tabletitle{\footnotesize{\parbox{1cm}{Pixel \\ (ROI)}}}}} &MAE &0.701 &\textbf{0.7} &0.713 &0.924 &\textbf{0.922} &0.938 \\
&PSNR &\textbf{21.69} &21.66 &21.58 &21.22 &21.23 &\textbf{21.24} \\
&SSIM &\textbf{0.514} &0.511 &0.509 &0.582 &\textbf{0.584} &0.579 \\\cmidrule{1-8}
\multirow{3}{*}{\rotatebox[origin=c]{90}{\tabletitle{\footnotesize{\parbox{1.3cm}{TI \\ metrics}}}}} &$\mathcal{ED}_{UR}$ &0.611 &0.612 &\textbf{0.493} &0.346 &0.404 &\textbf{0.334} \\
&$\mathcal{ED}_{R}$ &0.174 &0.178 &\textbf{0.122} &0.164 &0.17 &\textbf{0.14} \\
&$\mathcal{CP}_{s}$ &0.629 &0.621 &\textbf{0.827} &0.62 &0.614 &\textbf{0.818} \\
\bottomrule
\end{tabular}
\vspace{0.2cm}
\caption{Comparative results among all models using the test sets. MAE, SSIM and PSNR are reported as spatial feature metric for real/generated pair of images in the test set at two levels: full images and ROIs. Additionally, the metrics based in the TI curve ($\mathcal{ED}_{UR}$, $\mathcal{ED}_{R}$ and $\mathcal{CP}_{s}$) are also reported.} \label{tab_comp1}
\end{table}

Finally, we evaluated the performance of the PAN model and compared it with the CE-PAN and TI-PAN using the scores described in section \ref{sec_ti_metrics}. Table \ref{tab_comp1} shows the comparative performance among all models in 1.5T and 3T images. 

In the case of 1.5T images, pixel metrics are very similar, and the best performances are divided between PAN (SSIM full image, PSNR and SSIM ROIs) and TI-PAN (MAE and PSNR full image). These results align well with the qualitative images shown previously. Similarly, in the case of the 3T images, CE-PAN obtains the best performance in most metrics, except PSNR in ROIs. Nevertheless, the margin of difference between metrics is not large, and most models achieve similar values, with the largest difference obtained in PSNR, at 0.3 and 0.1 for 3T and 1.5T respectively. However, in the case of the TI metrics, significant differences were observed since TI-PAN obtains the best scores compared to PAN and CE-PAN. In the case of annotated ROIs extracted from tissue, the TI-PAN increases the estimation of CE intensity by $0.05\%$ in the case of 1.5T and $0.03\%$ in the case of 3T. Although the increase is not large, it might be crucial in the diagnosis of patients, as it might represent the difference between the different types of curves within the tissue.

A similar behavior is observed for the unannotated regions, where an increase in CE intensity estimation is noted for 1.5T images ($0.11\%$) and 3T images ($0.07\%$). This result is relevant because it demonstrates the ability of our model to estimate the intensity of CE, regardless of whether an anomaly is present in the tissue or not. This also indicates that the TI-PAN model does not rely on the presence of a tumor but rather leverages the biological behavior of the CA in the tissue. Finally, a large margin is reported in the $E_{ps}$ score, where an increase of $0.20\%$ is observed for both 1.5T and 3T images. The increase in performance is attributed to the TI-Norm and TI-Loss together, since they exploit the TI curve to synthesize the images.

\section{Discussion}

In breast DCE-MRI, late enhancement images, which can be acquired 8 or 10 minutes after the injection of the contrast agent, provide relevant information to distinguish benign from probably malignant lesions ~\cite{Rizzo2021Preoperative}. However, acquiring these images has the drawback of additional acquisition time and costs. Although alternatives such as abbreviated protocols have gained attention in clinical scenarios, concerns about their accuracy persist due to the omission of the late contrast enhancement phase, critical for an accurate diagnosis ~\cite{sundareswaran2020role,ijms77906,liu2023voxel}. Therefore, the development of alternatives for estimating the late contrast enhancement response is a valuable tool for moving towards their applicability in clinical practice. 

In this work, we propose a pipeline that uses a deep learning-based generative model to predict the late response of a DCE-MRI from its early counterpart, which is optimized to replicate the biological behavior of the contrast agent by introducing the Time-Intensity loss function. The pipeline leverages the behavior of contrast agents at multiple levels, demonstrating its ability to reliably outperform the synthesis of clinically relevant information while preserving the visual properties of the images when compared to existing models. 

To our knowledge, there are no previous works that have addressed the problem of synthesis or generation of late-enhanced images for breast DCE-MRI. However, some works that attempt to generate contrast-enhanced images have been reported. Those works have demonstrated adequate performance in generating realistic images with acceptable quality from a low-level feature perspective. Nonetheless, no evaluation of the ability to reproduce the behavior of the contrast agent pattern has been considered, which is a key factor in determining the feasibility of using it in clinical practice. The results presented in this work demonstrate that image quality metrics are not completely suitable for optimizing clinical information beyond visual similarity. Instead, task-specific types of functions, such as the proposed TI-Loss, could help achieve clinically interpretable results. 

The performance of the synthesis was evaluated in two ways: using established pixel-wise metrics and assessing the ability to replicate clinically relevant information (TI curve).  From the pixel-wise metrics perspective, experimental results showed sufficient realism and comparable performance in both the full image setting and selected regions (radiological findings and unannotated regions). On the other hand, from the clinical measures point of view, the proposed CA pattern ($\mathcal{CP}_{s}$) and the average enhancement differences ($\mathcal{ED}$) scores revealed significant differences that could seriously affect diagnostic decisions. In this case, the TI-PAN significantly outperformed other models, demonstrating that the proposed pipeline is able to maintain image quality during the synthesis process while replicating accurately the clinical performance of the model, preserving the full interpretability of the diagnostic information. 

Image normalization is a standard stage in any image synthesis process. We evaluated multiple normalization techniques commonly used in the field of medical image synthesis and found that the most common normalization approaches (i.e. Min-max and Z-score) significantly modify the behavior of the TI curve, distorting its objective interpretation. For this reason, we introduced the time-intensity normalization (TI-norm) strategy that preserves the diagnostic information required for physician interpretation. Our results in enhanced selected regions showed how the proposed TI-norm preserves the information in the TI curve beyond the visual cues.

This work differs from others because we propose an entire pipeline in which all stages are carefully intervened to preserve important information for the clinical scenario, rather than focusing solely on a single component, model, or function to process the data. Experimental results are promising for the implementation of late contrast-enhanced synthesis in a clinical setting. However, it must be evaluated through a controlled clinical study to determine the validity of the diagnostic information in a real assessment, particularly under varying acquisition parameters, variability among clinicians, and inter-reader interpretation, etc. In particular, performance with different contrast agents and dosages must be evaluated, considering that the pipeline is grounded in the qualification of the CA response and the TI-curve. These variations can cause differences in visualization and may alter the training process. This type of study is beyond the scope of this work and will be considered as a future line of research.

\section{Conclusion}

This work presents a comprehensive pipeline for synthesizing late-phase contrast-enhanced breast MR images from the early-phase counterpart. Our approach leverages the temporal behavior of contrast agents through the introduction of the Time Intensity Loss (TI-loss) function and a novel normalization strategy (TI-norm), which together guide the training of a generative model to accurately replicate clinically relevant information while preserving spatial features across the entire image, outperforming models optimized solely for pixel-wise quality. The synthesized images maintained diagnostic utility comparable to real late-phase images, as evidenced by the proposed evaluation metrics. Experimental results demonstrated that established pixel-based metrics do not guarantee the prevalence of contrast agent behavior, which is critical in clinical applications. 

By reducing the need for late-phase acquisitions, this pipeline has the potential to shorten scanning times, improve patient comfort, and alleviate cost and availability constraints associated with conventional breast DCE-MRI protocols. This advancement brings generative models closer to practical implementation in clinical scenarios, enhancing efficiency in breast cancer imaging without compromising diagnostic accuracy.

Future work must perform clinical validation through collaboration with healthcare professionals, which will be essential to ensure the method's efficacy and safety in real-world settings.

\section*{Ethics statement}

This study exclusively uses publicly available datasets, which are fully anonymized and do not contain identifiable personal or sensitive information. Datasets are accessible to the public and do not involve any interaction with human subjects.

\section*{Acknowledgements}

This work was supported by Minciencias, Colombia under the ``programa de becas de excelencia doctoral del bicentenario'', BPIN 2021000100029 and the Instituto Tecnológico Metropolitano.




\nocite{*} 

\end{document}